\documentclass[conference]{IEEEtran}
\IEEEoverridecommandlockouts
\usepackage{cite}
\usepackage[colorlinks,
            linkcolor=red,
            anchorcolor=black,
            citecolor=red]{hyperref}
\usepackage{amsmath,amssymb,amsfonts,mathrsfs,bbm }
\usepackage{algorithmic}
\usepackage{graphicx}
\usepackage{textcomp}
\usepackage{xcolor}
\usepackage{multirow,booktabs}
\def\BibTeX{{\rm B\kern-.05em{\sc i\kern-.025em b}\kern-.08em
    T\kern-.1667em\lower.7ex\hbox{E}\kern-.125emX}}
    
\DeclareRobustCommand*{\IEEEauthorrefmark}[1]{%
    \raisebox{0pt}[0pt][0pt]{\textsuperscript{\footnotesize\ensuremath{#1}}}}
\begin{document}

\title{HD2Reg: Hierarchical Descriptors and Detectors for Point Cloud Registration}
% \thanks{Identify applicable funding agency here. If none, delete this.}
% }
\author{
\IEEEauthorblockN{
Canhui Tang\IEEEauthorrefmark{1},
Yiheng Li\IEEEauthorrefmark{1},
Shaoyi Du\IEEEauthorrefmark{1,2,*},
Guofa Wang\IEEEauthorrefmark{3}, and
Zhiqiang Tian\IEEEauthorrefmark{3}}
\IEEEauthorblockA{\IEEEauthorrefmark{1}National Key Laboratory of Human-Machine Hybrid Augmented Intelligence, \\ National Engineering Research Center for Visual Information and Applications, \\and Institute of Artificial Intelligence and Robotics, Xi’an Jiaotong University, Xi’an, China}
\IEEEauthorblockA{\IEEEauthorrefmark{2}Shunan Academy of Artificial Intelligence, Ningbo, Zhejiang 315000, P.R. China}
\IEEEauthorblockA{\IEEEauthorrefmark{3}School of Software Engineering, Xi’an Jiaotong University, Xi’an, China}
\IEEEauthorblockA{\IEEEauthorrefmark{*}Corresponding author, Email: dushaoyi@gmail.com}}
% \author{\IEEEauthorblockN{1\textsuperscript{st} Canhui Tang}
% \IEEEauthorblockA{\textit{Institute of Artificial Intelligence and Robotics} \\
% \textit{Xi’an Jiaotong University
% }\\
% Xi’an, China\\
% 2193512014@stu.xjtu.edu.cn}
% \and
% \IEEEauthorblockN{2\textsuperscript{nd} Yiheng Li}
% \IEEEauthorblockA{\textit{Institute of Artificial Intelligence and Robotics} \\
% \textit{Xi’an Jiaotong University}\\
% Xi’an, China \\
% 1583898182@stu.xjtu.edu.cn}
% \and
% \IEEEauthorblockN{3\textsuperscript{rd} Shaoyi Du}
% \IEEEauthorblockA{\textit{Institute of Artificial Intelligence and Robotics} \\
% \textit{Xi’an Jiaotong University}\\
% Xi’an, China \\
% dushaoyi@gmail.com}
% \and
% \IEEEauthorblockN{4\textsuperscript{th} Guofa Wang}
% \IEEEauthorblockA{\textit{School of Software Engineering} \\
% \textit{Xi’an Jiaotong University}\\
% Xi’an, China \\
% wangguofa@stu.xjtu.edu.cn}
% \and
% \IEEEauthorblockN{5\textsuperscript{th} Zhiqiang Tian}
% \IEEEauthorblockA{\textit{School of Software Engineering} \\
% \textit{Xi’an Jiaotong University}\\
% Xi’an, China  \\
% zhiqiangtian@xjtu.edu.cn}
% }
% \author{\IEEEauthorblockN{Anonymous Authors}}

\maketitle

\begin{abstract}
Feature Descriptors and Detectors are two main components of feature-based point cloud registration.
However, little attention has been drawn to the explicit representation of local and global semantics in the learning of descriptors and detectors.
In this paper, we present a framework that explicitly extracts dual-level descriptors and detectors and 
performs coarse-to-fine matching with them.  First, to explicitly learn local and global semantics, we propose a hierarchical contrastive learning strategy, training the robust matching ability of high-level descriptors, and refining the local feature space using low-level descriptors. Furthermore, 
we propose to learn dual-level saliency maps that extract two groups of keypoints in two different senses. To overcome the weak supervision of binary matchability labels, we propose a ranking strategy to label the significance ranking of keypoints, and thus provide more fine-grained supervision signals. 
 Finally, we propose a global-to-local matching scheme to obtain robust and accurate correspondences by leveraging the complementary dual-level features.
 Quantitative experiments on 3DMatch and KITTI odometry datasets show that our method achieves robust and accurate point cloud registration and outperforms recent keypoint-based methods. \href{https://github.com/Hui-design/HD2Reg}{[code release]}
% by leveraging the complementary dual-level features.
% The high-level keypoint features are employed for global matching, and then low-level keypoint features are explicitly used for local matching within the constraints of coarse correspondences. 
\end{abstract}


\section{Introduction}
Point cloud registration refers to the problem of finding the optimal transformation that aligns two point clouds. With the development of deep learning, point cloud registration has shifted from raw data-based \cite{ICP,CPD} to feature-based approaches \cite{Predator,FCGF}, where descriptors and detectors are the two basic components. Descriptors are used to represent points' geometric and semantic information, and detectors are used to detect repeatable and reliable keypoints \cite{R2D2}.
% two folds: 1) improving the efficiency of registration by picking up a subset of point cloud; 2) improving registration performance by sampling matchable points. 

Traditional methods usually  extract low-level descriptors and detectors. They often employ handcrafted operators such as Harris \cite{Harris} and SIFT \cite{SIFT} to detect local keypoints, and then use handcrafted descriptors to characterize the patches around the keypoints. Such low-level descriptors remain local geometric details, and low-level detectors usually have good localization accuracy. However, they are less robust and not necessarily distinguishable, for they are easily affected by noise and repeated textures. 

Learning-based methods tend to extract high-level descriptors and detectors. Owing to the powerful representation ability of deep networks, high-level features learned from deeper layers are more robust for matching. Global distinctive descriptors, such as FCGF\cite{FCGF}, D3Feat\cite{D3Feat} and Predator\cite{Predator} show excellent ability to perform global registration, and high-level detectors\cite{R2D2,D2Net,D3Feat,Predator} focus on finding repeatable and matchable keypoints. However, high-level features usually lose local geometric details due to their abstraction.

% Using Unet-like network as backbone\cite{D3Feat, Predator} that fuses the
% dual-level features by skip connection may release the problem, yet still results in the insufficient representation of the global and local semantics.

% and detection based on high-level information leads to less accurate localization. 

% Owing to the powerful representation ability of deep networks and the expansion of receptive fields, high-level features learned from deeper layers hold more robust global matchability, i.e. they are less susceptible to noise or local repetitive textures.
% 

% \begin{figure}[t]
% \centering
% \includegraphics[width=1\columnwidth]{saliency.pdf} % Reduce the figure size so that it is slightly narrower than the column. Don't use precise values for figure width.This setup will avoid overfull boxes.
% \caption{An example to show the complementarity dual-level saliency map(detectors).  The local saliency map focuses on
% locations with local salient feature descriptors, which are not neccesarily matchable, such as corners and edges.
% Global saliency map focuses on points with globally matchable feature descriptors, which is more abstract and less accurate.}
% \label{fig1}
% \end{figure}

In a single network, the features extracted from the shallower layers are low-level, while those extracted from deeper layers are high-level. Prevalent point cloud registration methods \cite{FCGF,D3Feat,Predator} adopts a fully convolutional framework and simply uses skip connection to fuse the dual-level features, which is a implicit way and may suffer from insufficient representation of both the global and local semantics. To overcome the problem, we introduce a framework to explicitly learn hierarchical descriptors and detectors with different objectives and perform global-to-local matching with the features.

First, we propose a hierarchical contrastive learning strategy, training the robust matching ability of  high-level descriptors, and refining the local feature space using the low-level descriptors. In particular, our network outputs two kinds of descriptors, with local and global receptive fields, respectively. During training, low-level descriptors are guided to be distinctive from their local neighborhoods while high-level descriptors are guided to be distinctive from their global neighborhoods.

Based on the explicit dual-level features, we further explore global and local keypoint detection. Given the matching results of the dual-level feature during training, we propose to learn dual-level saliency maps. Local saliency map focuses on locations with local salient features, such as corners and edges. The global saliency map focuses on points with global matching ability, which is more abstract and less accurate. In addition, we propose a ranking strategy to label the significance ranking of keypoints, providing more fine-grained supervision signals compared with original binary matchability labels\cite{Predator}. 
% In particular, we propose a ranking strategy to label the significance ranking of global keypoints and local keypoints, respectively. The strategy provides more fine-grained supervision signals compared with original binary matchability labels. 

Based on the hierarchical descriptors and detectors, we propose a coarse-to-fine matching scheme by leveraging the complementarity of dual-level features. For global matching, high-level detectors and descriptors are employed consecutively to detect keypoints and feature matching. For local matching, low-level features are matched in local cells around each pair of correspondence.
Experiments on 3DMatch and KITTI odometry
datasets show that our method achieves robust and accurate point cloud registration and outperforms recent keypoint-based methods.

To summarize, our contribution are three-fold:
\begin{itemize}
    \item We propose a framework to explicitly learn and exploit hierarchical descriptors and detectors, achieving both the robustness and accuracy of point cloud registration.
   \item We propose a hierarchical contrastive learning strategy that trains the robust matching ability of  high-level descriptors and refines the local feature space using low-level descriptors.
    \item We propose to learn
    dual-level saliency maps that extract two groups of keypoints in two different senses. To overcome the weak supervision signals of original binary matchability labels, we propose a ranking strategy to label the saliency ranking of keypoints, providing more fine-grained supervision signals.
\end{itemize}

\section{Related Work}
Feature-based registration relies on the features that contain global or local semantics. Traditional methods usually extract low-level features, while learning-based methods tend to extract high-level features.

\textbf{Low-level descriptors and detectors.}
Early work uses hand-crafted 3D feature descriptors that can characterize local geometry. USC \cite{usc} uses covariance matrices of point pairs, SHOT \cite{shot} creates a 3D histogram of normal vectors, PFH \cite{pfh} and FPFH \cite{fpfh} build an oriented histogram using pairwise geometric properties. 
Two traditional Feature detectors are Harris \cite{Harris} and SIFT \cite{SIFT}, and their 3D version \cite{Harris3D,SIFT3D}. Harris measures the intensity of grayscale change of pixels and then adopts non-maximum suppression to find the local maxima. SIFT searches keypoints over all scales and image locations, with the help of the Difference of Gaussian(DoG) operator. These methods focus on local geometry, which is accurate but not robust to noise and repetitive patterns.

%  These methods are based on low-level feature maps, which have limited receptive fields range and thus are unable to distinguish repeating textures.  However, low-level descriptors remain local
% geometric details, and low-level detectors usually have good
% localization accuracy.

\textbf{High-level descriptors and detectors. }
Global features extracted from the deep layer of networks tends to be more robust but less accurate. Ding et al.\cite{GL} detection salient locations of point cloud  using local and global feature fusion. FCGF \cite{FCGF} extracts fully-convolutional geometric features based on a ResUnet architecture. D3Feat \cite{D3Feat} is inspired by D2Net and supervises meaningful saliency scores based on on-the-fly matching results during training, explicitly enforcing hand-crafted patterns in the descriptors.  Predator \cite{Predator} additionally designs a overlap detector and a self-supervised matchability detector. For the latter, it treats matchability prediction as a binary classification problem and also uses the matching results of current features as pseudo-labels of keypoint during training. However, high-level features usually lose local geometric details due to their abstraction. Fusing the dual-level features supervised by a single loss may release the
problem, yet still results in the insufficient representation of the global and local semantics.

\section{Method}
\begin{figure*}[t]
\centering
\includegraphics[width=0.95\textwidth]{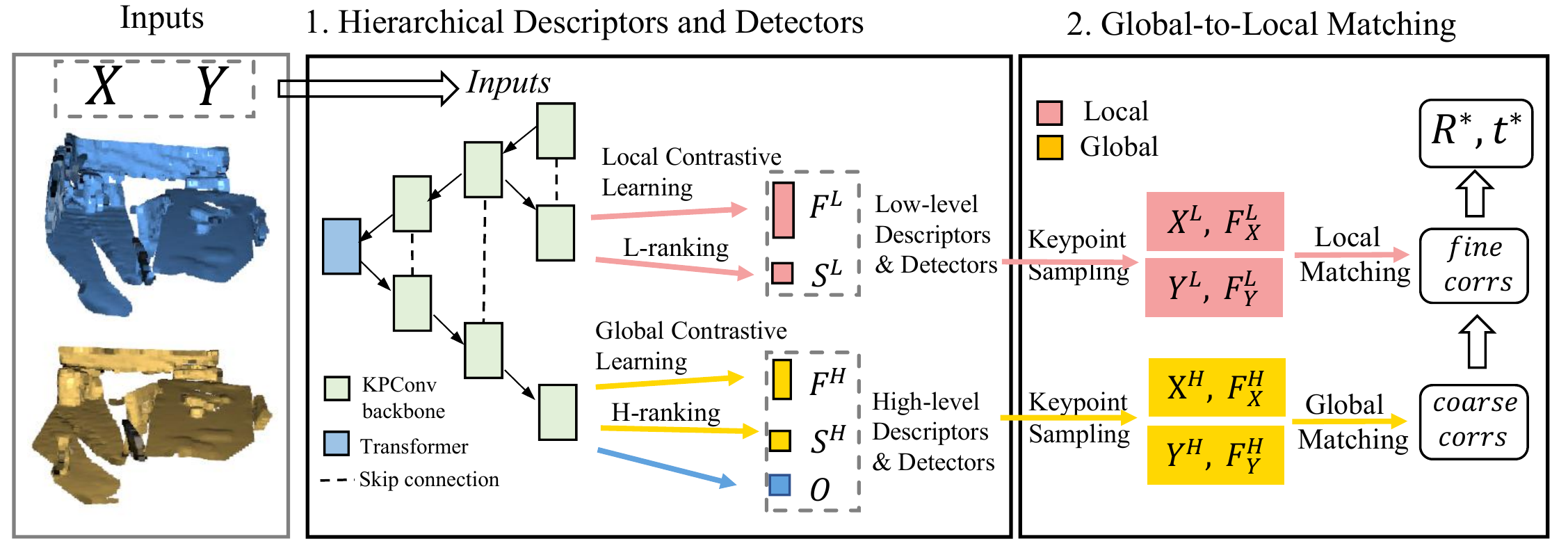} % Reduce the figure size so that it is slightly narrower than the column.
\caption{Overview of our method. (I) Hierarchical Feature Descriptors and Detectors. The module first extracts low-level descriptors $F^{L}$ (pink) and high-level descriptors $F^{H}$ (yellow). In particular, a hierarchical contrastive learning strategy is proposed to supervise local and global semantics, respectively. Based on the dual-level features, dual-level saliency scores $S^{L}$ and $S^{H}$ are predicted for keypoint detection, where a ranking strategy is proposed to provide more fine-grained supervision. (II) Global-to-local Matching. The sampled high-level keypoint locations $X^H, Y^H$ and descriptors $F_X^H, F_Y^H$ are employed for global matching to obtain coarse correspondences, and then the low-level locations $X^L, Y^L$ and descriptors $F_X^L, F_Y^L$ are explicitly used for fine registration within the constraints of coarse
correspondences.}
\label{fig1}
\end{figure*}

The proposed approach 
explicitly extracts and exploits hierarchical descriptors and detectors for the purpose of robust and accurate point cloud registration.  As illustrated in Fig. \ref{fig1}, our model predicts dual-level features descriptors and detectors, especially with hierarchical contrastive learning for descriptors (\ref{desc}), and keypoint ranking for detectors (\ref{det}). In addition, a global-to-local matching scheme (\ref{glm}) is proposed by leveraging the complementary
dual-level features.

% Our model starts from a feature extractor, which generates dense low-level and high-level descriptors. The low-level geometric descriptors $F^{low}_{X,Y}$ are learned from shallow layers, while the high-level semantic descriptors $F^{high}_{X,Y}$ are learned from deeper layers. In addition, two lightweight MLP heads are employed to predict dual-level scores $S^{low}_{X,Y}$ and $S^{high}_{X,Y}$ for matchability detection, guided by two self-supervised matchability labels. Furthermore, the high-level features are employed for global matching to obtain coarse correspondences, and then the low-level features are used for fine registration in an iteration-free manner.

\subsection{Hierarchical Contrastive Learning for Descriptors}\label{desc}
Feature-based registration relies on feature descriptors that contain local and global semantics. Our architecture starts from a feature extractor that aims to explicitly capture local and global semantics.

In particular, our network outputs two kinds of descriptors, with local and global receptive fields.
To learn features from point cloud data, KPConv-FCN\cite{Predator, D3Feat} backbone is adopted to predict pointwise feature descriptors. Then an additional upsampling block and a skip connection are introduced to construct a sub-network for extracting local features. 
% The re-designed feature extractor not only allows parameter sharing of local and global features, but also maintains both features at the point-wise resolution. 
% Thanks to the explicit learning of low-level features, high-level features do not necessarily maintain local details, so we cancel the skip connections between local and high-level features.
The low-level descriptors are learned from shallow layers, maintaining geometric details but lack of semantic information. In contrast, the high-level descriptors are learned from deeper layers with multiple downsampling and upsampling, thus having rich geometric information but losing some geometric details. In addition, considering partial overlapping is a tackle problem for point cloud registration, we simply follow \cite{Predator} and introduce an overlap module and predict overlap scores $O$ guided by an overlap loss to predict the overlapping region. 

\begin{figure}[t]
\centering
\includegraphics[width=0.48\textwidth]{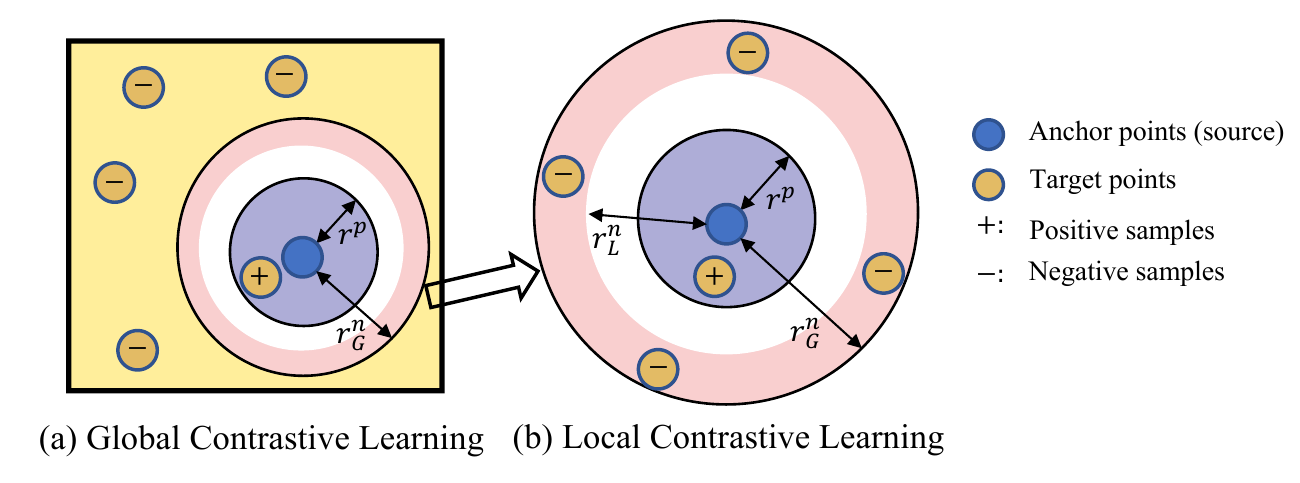} % Reduce the figure size so that it is slightly narrower than the column.
\caption{Hierarchical contrastive learning for descriptors.}
\label{fig2}
\end{figure}

\textbf{Hierarchical contrastive learning. } We propose a hierarchical contrastive learning strategy that trains robust matching ability in local and global semantics, respectively. The difference between the two supervisory signals lies in the selection of negative samples.
As shown in Fig. \ref{fig2} (a), for the high-level feature descriptors, a positive radius $r^{p}$ and a global negative radius $r_G^{n}$ are set to select positive samples and negative samples. Aligning the source point set to the target point set, and choosing some source points as anchor points, those within $r^{p}$ around the anchor points are regarded as positive samples (true correspondences), and others negative samples (false correspondences).
In the local neighborhood of anchor points, a local negative radius $r_L^{n}$ is introduced to further refine the local feature distinctiveness. 
As shown in Fig. \ref{fig2} (b), those within $r^{p}$ around the anchor points are regarded as positive samples, and those between $r_L^{n}$ and $r_G^{n}$ are negative samples. 

Based on the different selection of negative samples, we adopt a variety of common triplet loss, circle loss \cite{circle}, which has been widely used in point cloud feature descriptor learning \cite{Predator,D3Feat}. Given a set of correspondences from ground truth, the circle loss in the global phase is computed from $n_p$ points randomly sampled: 
\begin{equation}
\begin{aligned}
  L_{F}^{H}& = \frac{1}{n_p}\sum_{i=1}^{n_p}\log[1+\sum_{j\in\epsilon_p}{ e^{\beta_{P,j}^H(d_{i,j}^H-\Delta_p)}}\cdot\sum_{k\in\epsilon_n^G} e^{\beta_{n,k}^H(\Delta_n-d^H_{i,k})}],\\\\
L_{F}^{L} &= \frac{1}{n_p}\sum_{i=1}^{n_p}\log[1+\sum_{j\in\epsilon_p}{ e^{\beta_{p,j}^L(d_{i,j}^L-\Delta_p)}}\cdot\sum_{k\in\epsilon_n^L} e^{\beta_{n,k}^L(\Delta_n-d^L_{i,k})}],
\end{aligned}
  \label{eq1}
\end{equation}
where the $\epsilon_n^G$ and $\epsilon_n^L$ denote global negative samples and local negative samples, respectively. More details about the circle loss for point cloud can be referred to Predator\cite{Predator}.

\subsection{Keypoint Ranking for Detectors}\label{det}
Due to the lack of point cloud keypoint annotation datasets, most methods \cite{D2Net,D3Feat,Predator} adopt a self-supervised way to learn keypoint detectors. Predator \cite{Predator} predicts high-level matchability scores guided by the matching results of high-level descriptors, which results in two problems: (a) the lack of local saliency scores and (b) the binary matchability labels only provide coarse-grained supervision signals. 
To address the two problems, our framework predicts both local and global matchability scores, and provides multi-level labels based on a ranking strategy.

During training, binary matchability labels $\tilde{m}^{L}$ and $\tilde{m}^{H}$ can be computed based on the matching results of the dual-level descriptors during training. As shown in Eq. \eqref{eq2}, the matchability labels can be calculated by indicating whether the feature distance between positive samples $d_{pos}$ can be smaller than the feature distance of the closest negative samples $d_{neg}$.
% \begin{equation}
%   \tilde{m}^{low}_x = \mathbbm{1} \{||f^{low}_{x}-\frac{1}{|N_{x}|}\sum_{x^{'}\in N_x}f^{low}_{x
%  ^{'}}||_2<\tau\},
%   \label{eq4}
% \end{equation}
% where the distance threshold $\tau$ is set as the medium of all distance values. For the globally pseudo labels, we follow Predator, and define matchable keypoint are those that can be successfully matched with the on-the-fly high-level feature descriptors. This can be expressed as a pair-wise similarity learning problem. Given a triplet consisting of an anchor point and its corresponding and non-corresponding points, their feature distances are computed separately as:
\begin{equation}
\begin{aligned}
  \tilde{m}^{H} &= \mathbbm{1} \{d^{H}_{pos}-d^{H}_{neg}<0\},\\
  \tilde{m}^{L} &= \mathbbm{1} \{d^{L}_{pos}-d^{L}_{neg}<0\},
\end{aligned}
  \label{eq2}
\end{equation}
where $\mathbbm{1}$ is the indicator function, $\tilde{m}^{H}$ denotes the global matchability labels, and $\tilde{m}^{L}$ denotes the global matchability labels.

% In addition, $d^{high}_{pos}$ denotes the high-level feature distance of the corresponding point pair, and $d^{high}_{neg}$ denotes that of the non-corresponding point pair. Since the triplets are randomly picked over the point cloud, it indicates that a point is global matchable when $d_{pos}(i)<d_{neg}(i)$.

\textbf{Dual keypoint rankings.}
% The ranking of points is important for finding more excellent points, yet it is less explored in the field of keypoint-based point cloud registration.
Although binary labels can already provide supervision signals to some degree, the binary division is very coarse-grained and thus cannot distinguish which points are better. To provide more fine-grained supervision signals, we propose a keypoint ranking strategy.

In our model, since high-level key points benefit robust matching and low-level keypoints represent accurate locations, points can be naturally classified into four levels: robust and accurate, robust but inaccurate, accurate but unrobust, unrobust and inaccurate. 
Which level is ranked higher is a question. We follow the intuition that the high-level matchability enjoys a higher rank in the high-level ranking, while the low-level matchability occupies a higher rank in the low-level phase.
Therefore, the ranking of points can be cast as a 2-bit binary number, where high-level ranking corresponds to "$\tilde{m}^{H}\tilde{m}^{L}$", while low-level rankings correspond to “$\tilde{m}^{L}\tilde{m}^{H}$”,  In detail, the ratings can be computed as:
\begin{equation}
\begin{aligned}
 \tilde{r}^{H} &= 2\tilde{m}^{H}+\tilde{m}^{L}, \quad
 \tilde{r}^{L} &= 2\tilde{m}^{L} + \tilde{m}^{H},
\end{aligned}
\label{eq3}
\end{equation}
where $\tilde{r}^{H},\tilde{r}^{L}$ $\in\{0,1,2,3\}$.

\begin{figure}[t]
\centering
\includegraphics[width=0.38\textwidth]{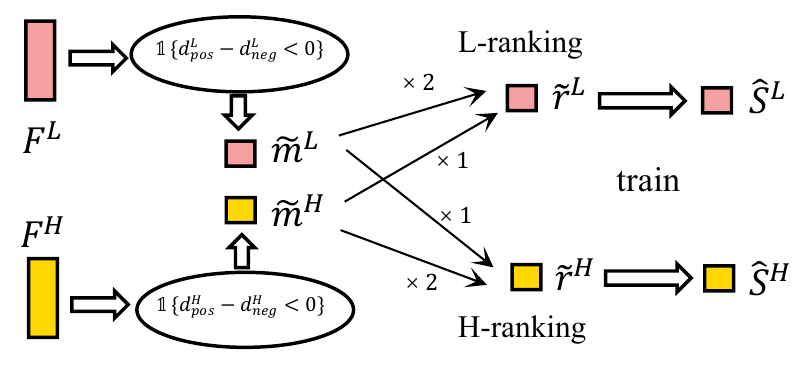} % Reduce the figure size so that it is slightly narrower than the column.
\caption{Dual keypoint ranking for detectors.}
\label{fig3}
\end{figure}

Furthermore, we propose dual rating losses guided by the high-level and low-level ratings. In details, the rankings $\tilde{r}$ are projected into target four-level scores: $\tilde{C}_3\textgreater \tilde{C}_2\textgreater \tilde{C}_1\textgreater \tilde{C}_0$, then the detection scores was supervised with two MSE losses:
\begin{equation}
\begin{aligned}
\mathcal{L}_{m}^{H} =& \frac{1}{M} \sum_{i=1}^{M}(\hat{S}_i^{H}-\tilde{C}_{r_i^{H}})^2 \\
\mathcal{L}_{m}^{L} =& \frac{1}{M} \sum_{i=1}^{M}(\hat{S}_i^{L}-\tilde{C}_{r_i^{L}})^2 \\
\end{aligned},
\label{eq10}
\end{equation}
where $\hat{S}_i^{H}$ is the predicted detection score and $M$ is the number of sampled points during training. The subscript $r_i^{H}$ controls the level of target score $\tilde{C}_{r_i^{H}}$,  and $r_i^{L}$ is the same. 

\subsection{Global-to-Local Matching}\label{glm}
High-level features are more robust for global matching, low-level features are more accurate for local feature matching. Aiming to leveraging the complementarity of the dual-level feautures, we propose a coarse-to-fine matching scheme.

For global matching, high-level detectors and descriptors are employed consecutively to detect keypoints and feature matching. In detail, coarse correspondences $ \{X,X^{corr}\}$ are solved by high-level global feature matching in the global phase using high-level keypoint features $F^{H}_X$ and $F^H_Y$.

% :
% \begin{equation}
% \begin{aligned}
% \{X, X^{corr}\} = GFM(F^{H}_X, F^{H}_Y),
% \end{aligned}
% \label{eq5}
% \end{equation}
% where $\{X,X^{corr}\}$ denotes the global corresponding point coordinates.

For local matching, low-level features are matched in local cells around each pair of the coarse correspondence. In particular, each node finds its neighborhood by radius searching in coordinate space. Given a group of points$\{x_i, y_i\}$ in the local cell for coarse correspondence $i$, fine-grained correspondences are established via low-level local feature matching. Afterward, all groups of fine correspondences are collected and an excellent subset of them is sampled by employing the low-level detectors. Finally, optimal transformation $T^*=\{R^*,t^*\}$ can be solved from weighted SVD \cite{svd}.
% \begin{equation}
% \begin{aligned}
% \{x_i, x_i^{corr}\} &= LFM(F^{L}_{x_i},F^{L}_{y_i})\\
% \{R,t\} &= SVD(\cup_{i=0}^{i<N_c} \{x_i, x_i^{corr}\})
% \end{aligned},
% \label{eq4}
% \end{equation}
% where $\{x_i,x_i^{corr}\}$ denotes the local corresponding point coordinates, and $N_c$ is the number of global correspondences.

% Compared with the refine registration approaches based on ICP \cite{ICP}, our fine registration based on low-level features can find the accurate corresponding points without multiple iterations. At the same time, the accuracy of the corresponding relationship is guaranteed under the constraint of high-level global matching.

% \subsection{Comparisons to related work}\label{sec2}
% \textbf{Ours vs. hand-crafted detectors \cite{Harris,SIFT}}
% Hand-crafted methods detect keypoints on low-level feature map, which corresponds to the low-level branch
% of our architecture. However, we also have an additional high-level branch for providing robust registration initials.\\
% \textbf{Ours vs. Predator\cite{Predator}}
% First, predator only extracts high-level descriptors and detector, while our method also utilizes a low-level branch to achieve accurate registration. Second, to improve the weak supervision signal for the matchability score in Predator[], a rating strategy and a rating loss are proposed in our work to give stronger supervision signals.\\

\subsection{Implementation Details}
Our method is implemented in pytorch and can  be trained on a single Tesla M40 with
Intel(R) Xeon(R) E5-2690 CPU with 128G RAM.
The overall objective function of our model comprises of dual-level descriptor losses, an overlap loss, and dual-level matchability losses:
\begin{equation}
 \mathcal{L}= \{\mathcal{L}_F ^{H}+\mathcal{L}_F^{L}\}+ \mathcal{L}_o + \{\mathcal{L}_m ^{H}+\mathcal{L}_m^{L}\},
% \end{aligned}
\label{eq5}
\end{equation}
where $\mathcal{L}_F$ denotes the descriptor loss, $\mathcal{L}_o$ denotes the overlap loss, and $\mathcal{L}_m$ denotes the matchability loss. Due to the problem of partial overlapping, we follow Predator \cite{Predator} and obtain detection scores by multiplying the matchability scores and overlap scores.

Since the supervision of matchability relies on the reliability of descriptors, we first pre-train the feature extractor with the descriptor losses and overlap loss for 20 epochs and then introduce the matching loss to train the two detection heads together. For hyperparameters about the circle loss and overlap loss, we follow the settings of Predator \cite{Predator}.
For the global-to-local matching, global feature matching is implemented by RANSAC, a robust correspondence estimator. In the local feature matching phase, the radius of neighborhood search for low-level matchability labels is set to 0.1m.  
For the training of matchability loss, 256 point pairs are randomly sampled, and target scores $\{\tilde{C}_3,\tilde{C}_2,\tilde{C}_1,\tilde{C}_0\}$ are set to $\{1.0,0.75,0.25,0.0\}$.   The optimizer is SGD with an initial learning rate of 0.005 and momentum of 0.98.

%  The number of encoders layers $L$ of the modified Unet backbone needs to be adjusted slightly with the point cloud scene. $L$ is set to 3 in 3DMatch, while it is set to 4 in KITTI odometry due to its larger scale. 
%

\section{Experiments}
Our model is evaluated on both indoor 3DMatch \cite{3DMatch}
and 3DLoMatch \cite{Predator} benchmarks, and outdoor KITTI odometry \cite{KITTI} benchmark.

\subsection{3DMatch $\&$ 3DLoMatch}
\textbf{Datasets.} 
3DMatch is an indoor dataset reconstructed from RGBD images and consists of point cloud data for 62 scenes. We follow the protocols \cite{3DMatch} to split training, validation, and testing datasets. The original testing datasets only contain point cloud pairs with \textgreater30$\%$ overlap. Predator \cite{Predator} proposes a harder dataset with low overlap, 3DLoMatch, which collects point cloud pairs with overlapping ratios between 10$\%$ and 30$\%$.

\textbf{Metrics.} Since the actual aim of point cloud registration is to recover the transformation between two fragments, our main metric consists of two parts: 1) Registration Recall, the fraction of scan pairs where the correct transformation is recovered, which measures the robustness of registration \cite{RR}. 2) Relative Rotation Error (RRE) and Relative Translation Error (RTE), the deviations from the ground truth pose, which measure the accuracy of registration \cite{Predator}.

\begin{table}[t]
  \centering
  \caption{Registration results on the 3DMatch and 3DLoMatch datasets.}
  \setlength\tabcolsep{2pt}
\scriptsize
    \begin{tabular}{cccccc|ccccc}
    \toprule
 & \multicolumn{5}{c|}{3DMatch} & \multicolumn{5}{c}{3DLoMatch} \\
    $\#$Samples & 5000&2500&1000&500&250$\,$&$\,$5000&2500&1000&500&250\\
		\midrule
		& \multicolumn{10}{c}{Registration Recall($\%$)$\uparrow$} \\
		\midrule
% 		FPFH+ICP[]&36&&&&$\,$&&&&&\\
		FCGF\cite{FCGF}&85.1&84.7&83.3&81.6&71.4$\,$&40.1&41.7&38.2&35.4&26.8\\	D3Feat\cite{D3Feat}&81.9&84.5&83.6&80.7&70.9$\,$&35.2&39.7&39.5&34.5&22.9\\	
	Predator\cite{Predator}&89.0&89.9&90.6&88.5&86.6$\,$&59.8&61.2&62.4&60.8&58.1\\
	% CoFiNet\cite{CoFiNet}&89.3&88.9&88.4&87.4&87.0&\textbf{67.5}&\textbf{66.2}&64.2&63.1&61.0\\

% Ours(H)&89.9&90.2&90.7&88.9&86.6$\,$&\textbf{63.0}&\textbf{62.8}&\textbf{65.6}&\textbf{64.0}&\textbf{60.5}\\
Ours&\textbf{90.4}&\textbf{90.6}&\textbf{91.0}&\textbf{90.2}&\textbf{89.9}$\,$&\bf{62.9}&\bf{62.9}&\textbf{64.7}&\textbf{64.1}&\bf{63.5}\\
\midrule
		& \multicolumn{10}{c}{RRE($^{\circ}$)$\downarrow$} \\
		\midrule
% 		FPFH+ICP[]&\textbf{1.461}&&&&$\,$&&&&&\\
		FCGF\cite{FCGF}&1.911&1.926&2.165&1.953&3.369\,&3.086&3.152&3.438&3.875&4.675\\	D3Feat\cite{D3Feat}&2.059&1.966&2.311&2.808&3.505\,&3.206&3.403&3.755&4.367&5.156\\	
	Predator\cite{Predator}&1.925&2.036&2.231&2.212&2.717$\,$&3.071&3.106&3.091&3.416&3.868\\
	% CoFiNet\cite{CoFiNet}&2.002&2.124&2.281&2.302&2.486\,&3.271&3.415&3.520&3.513&3.748\\
% Ours(H)&89.9&90.2&90.7&88.9&86.6$\,$&\textbf{63.0}&\textbf{62.8}&\textbf{65.6}&\textbf{64.0}&\textbf{60.5}\\
Ours&\textbf{1.811}&\textbf{1.824}&\textbf{1.832}&\textbf{1.845}&\textbf{1.940}$\,$&\textbf{3.032}&\textbf{3.033}&\textbf{3.027}&\textbf{3.015}&\bf{3.234}\\
\midrule
		& \multicolumn{10}{c}{RTE($m$)$\downarrow$}\\
		\midrule
% 		FPFH+ICP[]&\textbf{0.049}&&&&$\,$&&&&&\\
	FCGF\cite{FCGF}&0.065&0.065&0.075&0.085&0.102\,&0.096&0.103&0.111&0.113&0.131\\	D3Feat\cite{D3Feat}&0.070&0.069&0.073&0.088&0.106\,&0.105&0.099&0.111&0.118&0.129\\
    Predator\cite{Predator}&0.066&0.069&0.071&0.073&0.084$\,$&0.096&0.095&0.092&0.096&1.017\\
	% CoFiNet\cite{CoFiNet}&0.064&0.063&0.069&0.070&0.074\,&0.090&0.095&0.096&0.099&0.107\\
% Ours(H)&89.9&90.2&90.7&88.9&86.6$\,$&\textbf{63.0}&\textbf{62.8}&\textbf{65.6}&\textbf{64.0}&\textbf{60.5}\\
Ours&\textbf{0.057}&\textbf{0.057}&\textbf{0.060}&\textbf{0.060}&\textbf{0.064}$\,$&\textbf{0.088}&\textbf{0.089}&\textbf{0.089}&\textbf{0.088}&\bf{0.093}\\
    \bottomrule
    \end{tabular}
     \label{tab1}
\end{table}
% 3) Feature Match Recall(FMR), defined as the fraction of pairs with >5$\%$ ”inlier” matches, can directly reflect the matchability of keypoints' descriptors and is also related to the joint capacity of descriptors and detectors. \\

Our method is compared with the recent keypoint-based methods: FCGF\cite{FCGF}, D3Feat \cite{D3Feat} and Predator \cite{Predator}. Since recent patch-based methods \cite{CoFiNet, GeoTransformer} are keypoint-free and different enough from our framework, we do not compare them.
For the strategy of sampling, we follow \cite{Predator} and sample points with probability proportional to the detector scores, where the scores are obtained by multiplying the matchability scores and overlap scores. The sampling number of keypoints for global matching varies from 5000 to 250, and the sampling number is fixed at 1000.

\textbf{Registration robustness and accuracy. }
To evaluate the registration robustness of our method, we first report the Registration Recall in Table \ref{tab1} (top). Our method outperforms recent keypoint-based methods under different overlap ratios and different keypoint sampling numbers, which demonstrate the registration robustness of our method. In particular, our method performs much better than other methods when the number of sampling is less than 1000, which makes our method potentially applicable to real-time tasks. 

We then report the RRE and RTE in Table \ref{tab1} ($2^{nd}$ and $3^{rd}$ rows) to demonstrate the registration accuracy of our method. They significantly outperforms other methods both in 3DMatch and 3DLoMatch, and shows lower error when the sampling numbers increase. The results show that our method achieves accurate registration.

% For our single-iteration fine transformation, we fix the sampling number to 2500, which makes our fine registration much more efficient than ICP. 
% For a better understanding of each module in 
% Our fine registration is based on low-level which is better for accurate and reliable correpondence, so it outperform ICP both in speed and accuracy. we adopt the implementation of ICP in[], and compare our fine registration with ICP registration, 

\begin{table}[htbp]
  \centering
   \caption{Performance of our dual-level descriptors and detectors. The sampling number of keypoints is set to 500.}
  \setlength\tabcolsep{2.5pt}
\scriptsize
    \begin{tabular}{cccc|ccc}
    \toprule
     & \multicolumn{3}{c|}{3DMatch} & \multicolumn{3}{c}{3DLoMatch} \\
Method&IR ($\%$)&FMR ($\%$)&Rep ($\%$)&IR ($\%$)&FMR ($\%$)&Rep ($\%$)\cr
    \midrule
    FCGF\cite{FCGF}&42.5&96.7&-&14.8&71.7&-\\D3Feat\cite{D3Feat}&41.5&94.1&51.2&14.6&66.7&22.8\\
Predator\cite{Predator}&54.1&96.3&80.8&27.5&75.7&59.1\\
    % CoFiNet\cite{CoFiNet}&52.2&\bf{98.2}&-&26.8&83.1&-\\
    Ours(high-level)&54.7&96.4&77.9&27.6&78.7&57.8\\
    Ours(low-level)&23.8&92.7&73.4&10.3&70.4&50.6\\
    Ours(global-to-local)&\textbf{69.6}&96.6&\textbf{91.3}&\textbf{45.6}&\textbf{83.5}&\textbf{71.9}\\
    \bottomrule
    \end{tabular}
  \label{tab2}
\end{table}

% \begin{figure*}[thbp]
% \centering
% \includegraphics[width=0.9\textwidth]{pipeline(6).pdf}
% \caption{Visual comparison of the baseline and our HD2Reg framework. Compared with the pipeline of Predator\cite{Predator} (a), our hierarchical descriptors and detectors (b) achieves more robust and accurate registration results. The 32-dimensional descriptors for the pair of point
% clouds are mapped to a scalar space using t-SNE \cite{van2008visualizing}and colorized with the Oranges color map. In addition, the small red spheres are the extracted keypoints sampled from the detector map. }
% \label{fig3}
% \end{figure*}

\textbf{Performance of descriptors and detectors.}
The registration performance depends on the joint capabilities of the descriptors and detectors, so we also report this with three metrics: 1) Inlier Ratio (IR): the fraction of correct correspondences obtained from feature matching \cite{FCGF}, 2) Feature Matching Recall (FMR): the fraction of pairs that have $\textgreater$5$\%$ ”inlier” matches \cite{FCGF}, 3) Repeatability (Rep): the fraction of repeatable keypoint locations detected by detectors, which reflects the individual performance of detectors. In the baselines, FCGF does not need to report Rep because it does not learn detectors to predict keypoints.

\begin{figure}[t]
\centering
\includegraphics[width=0.45\textwidth]{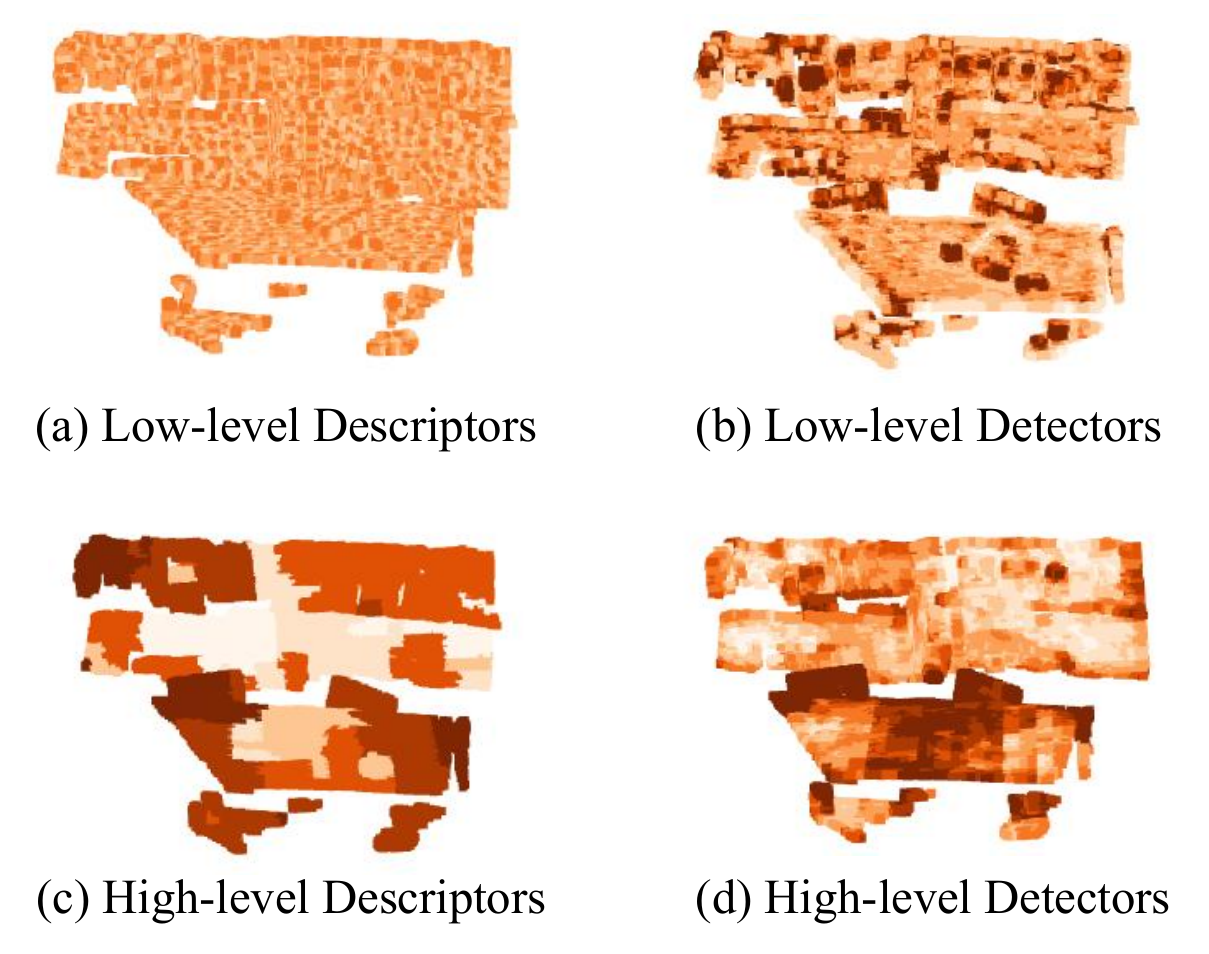} % Reduce the figure size so that it is slightly narrower than the column.
\caption{Qualitative results of the dual-level descriptors and detectors. Low-level ones focus on local details while high-level ones focus on global distinctiveness.}
\label{fig4}
\end{figure}

Since our high-level Feature branch adopts a similar backbone as Predator \cite{Predator}, the performance of our high-level descriptor and detectors is very close to it. When the low-level features are evaluated without the prior constraints, they achieve low matchability but still high repeatability. We then evaluate the joint leveraging of dual-level features (global-to-local). By employing coarse global correspondences and local refinement, our method yields the best performance with IR of 69.6 $\%$ and Rep of 91.3 $\%$ (see Table \ref{tab2}). Thanks to the reliable correspondences, the transformation parameters can be solved with SVD as described above. Qualitative results of our dual-level descriptors and detectors can been seen from Fig. \ref{fig4}.

\begin{table}[htbp]
  \centering
\caption{Ablation studies of the dual-level ranking strategy, where we analyze the impact of the low-level rankings
and high-level rankings. The sampling number of keypoints is set to 500.}
  \scriptsize
\setlength\tabcolsep{5pt}
    \begin{tabular}{cc|ccc|ccc}
    \toprule
high-level&low-level &\multicolumn{3}{c|}{\quad\quad 3DMatch\quad\quad}&\multicolumn{3}{c}{\quad\quad 3DLoMatch\quad\quad}\\
 rankings&rankings&RR&RRE&RTE&RR&RRE&RTE\\
\midrule
&&88.9&1.996&0.071&63.0&3.211&0.095\\
\checkmark&&90.0&1.832&0.071&64.0&3.100&0.089\\
&\checkmark&89.6&1.856&0.075&63.8&3.125&0.092\\
\checkmark&\checkmark&\bf{90.2}&\bf{1.845}&\bf{0.060}&\bf{64.1}&\bf{3.015}&\bf{0.080}\\
    \bottomrule
    \end{tabular}
  \label{tab3}
\end{table}
% \begin{table}[htbp]
%   \centering
% \setlength\tabcolsep{1pt}
%   \caption{Ablation study of our network architecture. }
%     \begin{tabular}{cccccc|ccc}
%     \toprule
% \multicolumn{2}{c}{\underline{\quad\quad DLF\quad\quad}}&\multicolumn{2}{c}{\underline{\quad\quad DLD \quad\quad}}&\multicolumn{2}{c|}{\underline{\quad\quad DLR\quad\quad}}&\multicolumn{3}{c|}{\underline{\quad\quad3DMatch\quad\quad}}\\
% HDesc&LDesc&HDet&LDet&HReg&LReg&RR&RRE&RTE\\
% \midrule
% \checkmark&&&&\checkmark&&83.7&2.394&0.076\\
% \checkmark&&\checkmark&&\checkmark&&90.4&2.165&0.069\\
% &\checkmark&&&&\checkmark&69.0&2.642&0.080\\
% &\checkmark&&\checkmark&&\checkmark&86.1&2.322&0.070\\
% \checkmark&\checkmark&\checkmark&\checkmark&\checkmark&\checkmark&\bf{91.3}&\bf{1.897}&\bf{0.060}\\
%     \bottomrule
%     \end{tabular}
%   \label{tab2}
% \end{table}

% \ref{fig3}, 
% \begin{figure}[htbp]
% \centering
% \subfigure[RRE]{
% \begin{minipage}[t]{0.5\linewidth}
% \centering
% \includegraphics[width=1in]{111.eps}
% %\caption{fig1}
% \end{minipage}%
% }%
% \subfigure[RTE]{
% \begin{minipage}[t]{0.5\linewidth}
% \centering
% \includegraphics[width=1in]{111.eps}
% %\caption{fig2}
% \end{minipage}%
% }%
% \centering
% \caption{pics}
% \end{figure}
\textbf{Ablation studies of the keypoint ranking strategy.} Another contribution of the proposed method is the supervision signals for detectors, where the dual keypoint rankings are almost unexplored in the field of point cloud registration. Here we analyze the impact of the low-level rankings and high-level rankings. We first remove the dual-level rankings and simply use the supervision of the original binary matchability losses then we only add the high-level rankings, the low-level rankings, and the combination of them. As shown in Table \ref{tab3}, the combination of dual-level rankings achieves best registration results. In addition, the high-level ranking has a more significant impact on increasing performance. One possible reason is that the raw binary matching labels provide weaker supervision, Furthermore, high-level rankings have a more significant impact on improving performance. One possible reason is that the raw binary matching labels provide weak supervision, as many points lying on the smooth plane are labeled as positive samples due to the overfitting of high-level features.

% \textbf{Influence of the number
% of iterations on our local refinemeant.} As mentioned above, our fine registration based
% on low-level features can find the accurate corresponding
% points without multiple iterations. At the same time, the ac-
% curacy of the corresponding relationship is guaranteed under
% the constraint of high-level global matching. As illustrated in Figure 

% \textbf{Qualitative results. }
% As shown in Figure \ref{fig3}, we first visualize the pipeline of our baseline, Predator \cite{Predator}. It can be seen from the score map (3rd column) that the inaccurate prediction of the overlapping area results in low inlier ratio, which leads to the failure of transformation.
% We then visualize the pipeline of our method. It first extracts dual-level feature descriptors and then generated dual-level score maps guided by two different losses. It can be seen that the low-level score map is good at finding local salient points, while the high-level score map is good at finding globally matchable points. Although our score map also cannot predict the overlapping area well, the keypoints we detected had higher localization accuracy (4th column), which facilitated accurate matching. At the same time, further refined the Local Reduce the relative error of the two frames. Finally, the estimated transformation is very close to the ground truth.

\begin{table}[htbp]
  \centering
\caption{Registration results on KITTI odometry}
  \scriptsize
\setlength\tabcolsep{12pt}
    \begin{tabular}{cccc}
    \toprule
Method & RTE[cm]&RRE[$^{\circ}$]&RR[\%]\\
		\midrule
		3DFeat-Net\cite{3DFeat-Net}&25.9&0.57&96.0\\
		FCGF\cite{FCGF}&9.5&0.30&96.6\\
		D3Feat\cite{D3Feat}&7.2&0.30&99.8\\
	Predator\cite{Predator}&6.8&0.27&99.8\\
		% CoFiNet\cite{CoFiNet}&8.2&0.41&99.8\\
		Ours&\bf{6.3}&\bf{0.27}&\bf{99.8}\\
    \bottomrule
    \end{tabular}
  \label{tab5}
\end{table}
\subsection{KITTI odometry }
\textbf{Dataset.}
KITTI \cite{KITTI} is a sparse outdoor LiDAR dataset, containing 11 sequences of outdoor driving scenarios. We follow \cite{Predator} and split datasets and refine the ground-truth transformation by ICP.

We follow \cite{Predator} to evaluate our method with three metrics: (1) Relative Rotation Error (RRE), the
geodesic distance between estimated and ground-truth rotation matrices, (2) Relative Translation Error (RTE), the Euclidean distance between estimated and ground-truth translation vectors, and (3) Registration Recall (RR), the fraction
of point cloud pairs whose RRE and RTE are both below certain thresholds (i.e., RRE$\textless$5$^\circ$and RTE$\textless$2m). On KITTI Odometry, HD2Reg is compared with 3DFeat-net \cite{3DFeat-Net}, FCGF \cite{FCGF}, D3Feat \cite{D3Feat}, PREDATOR \cite{Predator}. Quantitative results can be found in Table \ref{tab5}, our model outperforms all other methods with 6.3 RTE and 0.27 RRE.

\section{Conclusion}
In this paper, we introduce a coarse-to-fine framework that explicitly learns dual-level feature descriptors and dual-level feature detectors for robust and accurate point cloud registration. First, we propose a hierarchical contrastive learning strategy
that trains the robust matching ability of high-level descriptors and refines the local feature space using low-level descriptors. Second, we propose to learn dual-level saliency maps that extract two groups of keypoints in two different senses. To overcome the weak supervision signals of original binary matchability labels, we propose a ranking strategy to label the saliency ranking of keypoints, providing more fine-grained supervision signals. Quantitative experiments on 3DMatch and KITTI odometry datasets show that our method achieves robust and accurate point cloud registration and outperforms recent keypoint-based methods.
% There are some directions in which our method could be extended. For example, we assume that the global matching can basically align the point clouds at present. However, in case the global matching is not successful, it is not easy for our model to recover the correct pose during the local matching stage.

\section*{Acknowledgment}
This work was supported by the National Key Research and Development Program of China under Grant No. 2020AAA0108100.
\bibliographystyle{IEEEtran}
% \bibliography{ref.bib}

\end{document}